\documentclass[runningheads]{llncs}
\usepackage{graphicx}
\usepackage{amsmath}        
\usepackage{amsfonts}       
\usepackage{tikz}
\usepackage{subcaption}
\usepackage{tcolorbox}
\usepackage{xcolor}
\usepackage{multirow,tabularx}

\usepackage{float}
\usepackage{mathtools}

\usepackage{textcomp}

\usepackage[warning]{numprint}
\npdecimalsign{.}
\nprounddigits{3}

\begin{document}

\title{Supervised Phrase-boundary Embeddings}
%
%
\author{Manni Singh \and
David Weston \and
Mark Levene}
\authorrunning{Manni et al.}
%
\institute{Department of Computer Science and Information Systems, \\ 
Birkbeck, University of London, London WC1E 7HX, U.K.\\
\email{\{manni,dweston,mark\}@dcs.bbk.ac.uk}}
\maketitle              
\begin{abstract}
We propose a new word embedding model, called SPhrase, that incorporates supervised phrase information. Our method modifies traditional word embeddings by ensuring that all target words in a phrase have exactly the same context. We demonstrate that including this information within  a context window produces  superior embeddings for both intrinsic evaluation tasks and downstream extrinsic tasks.

\end{abstract}
\section{Introduction}

Word embeddings represent words with multidimensional vectors that are used in  various models for applications such as, named entity recognition \cite{ner_lstm}, query expansion \cite{query_expansion}, and sentiment analysis \cite{sentiment_analysis}. These embeddings are usually generated from a huge corpus with unsupervised learning models \cite{w2v,glove,fasttext,deps,lexvec}. These models are based on describing target words by their neighbouring words which are also considered as contexts. The selection of these context words is generally linear (i.e. $n$ words surrounding the target). Alternatively, arbitrary context words were used in \cite{deps} where context selection is based on the syntactic dependencies to the target word.

These models treat words as lexical units and create a context window surrounding a target word. This approach can be problematic when the context window for a target word contains only part of a phrase. For example, consider a scenario where a target word is close to (and to the right of) the named entity ``George W. Bush''  but the context window only retains the  word ``George''.  Clearly this will  generate ambiguity as the independent word ``George'' may refer another person (George Washington), location (George Street, Oxford) or a music band (George).
To deal with the issue described above, \cite{w2v2} used a data-driven approach to identify and treat these phrases as individual tokens. While this technique may learn a phrase representation  it cannot learn a representation of the individual words  that comprise the phrase.

In our approach we obtain phrase information directly from Wikipedia. Terms from Wikipedia articles are formatted as hyperlinks to relevant articles. In a related method \cite{wikipedia} these terms are extracted as named entities. This paper interprets these terms as phrases. By using Wikipedia for phrase information (unlike \cite{deps}) we avoid needing additional grammatical information. This also gives us the potential to generate multi-lingual embeddings, although we do not pursue this here.  

In this work, we are using phrase boundary information to generate word embedding in a non-compositional manner rather than a phrase embedding. We consider each of the words in the phrase as a part of the unit, where a unit can either be single word (i.e. not a link in the Wikipedia) or otherwise a bag of words. The embeddings are then learned for each of the unit members by considering surrounding units in the context.    

In the following section we present related work in this domain, Section \ref{model} presents our model and in Sections \ref{implimementation} to \ref{experiment} we give details of the implementation and the experiments.

\section{Related Work}
\label{related}

Word representations can be obtained from a language model where the goal is to predict a future word based on some previously observed information such as, a sentence,  a sequence, or a phrase. For this task, various models can be utilised including: joint probabilities of observation that may include the Markov assumption.  Under this assumption, we may say that the immediate future is independent of the entire past given the present. N-gram language models \cite{ngramHuge} use this assumption to predict token(s)  using the previous $N-1$ tokens \cite{martinJurafsky}. This can be constructed efficiently for very large datasets using neural network based language modelling (NNLM) \cite{bengio2003neural}. 

The NNLM of \cite{bengio2003neural} used a non-linear hidden layer between the input and output layers. A simpler network named the log bi-linear model was introduced in \cite{three} by dropping the hidden layer between input and output layer. Instead of the hidden layer, context vectors were summed and projected to the output layer. This model was later used by \cite{w2v} and named CBOW (Continuous Bag-of-words model), with a symmetric context (i.e. context words on both sides of the target word).

In addition, the Skip-gram model, was introduced in this work by reversing CBOW to predict context from the target word. Given a context range $c$ and target word $w_t$ the objective is to maximise the average log probability,

\begin{equation*}
\sum_{-c\leq j\leq c} \log p(w_{t+j}|w_t)
\end{equation*}

\noindent The model defines $p(w_{t+j}|w_t)$ using the softmax function, 

\begin{equation*}
p(w_O|w_I) = \frac{\exp\left({v'_{w_O}}^\top v_{w_I}\right)}
{\sum_{w=1}^{W}\exp\left({v'_w}^\top v_{w_I}\right)}
\end{equation*}

where $v_w$ and $v'_w$ are the ``input'' and ``output'' vector representations
of $w$, and $W$ is the number of words in the vocabulary. However, due to the large vocabulary, the computation becomes impractical. Thus, Noise Contrastive Estimation (NCE) \cite{nce} was used that performs the same operation by sampling a very small amount of words $k$ from the vocabulary as noise.

A similar technique is called Candidate Sampling \cite{candidate_sampling} that combines noise samples with the true class, denoted as the set $\mathcal{S}$, with the objective to predict the true class from it, where $Y$ is a set of true classes. Embeddings are scored as, 

\begin{equation*}
\hat{Y}_s = (X_s*W_s+b_s) - \log(E(s)).
\end{equation*}

Where $X_s$ is a vector (embedding) corresponding to a word $s \in \mathcal{S}$, $W_s$ is the corresponding weight, $b_s$ is the bias, and $\mathbb{E}(s)$ is the expectation for s. Each score is approximated to a probability using the softmax function,

\begin{equation*}
Softmax(\hat{Y}_s) =\dfrac{\exp{\hat{Y}_s}}{\sum_{s' \in S}{\exp{\hat{Y}_{s'}}}}.
\end{equation*}

In addition to words, phrases may also be considered. In \cite{w2v}, the words comprising a phrase were joined using the delimiter '\_' between them, and their joint embedding was learned. This scheme is called non-compositional embedding \cite{LearningCM,AdaptiveJL}. Alternatively, compositional embeddings \cite{AdaptiveJL} are generated by merging word embeddings of phrase components using a composition function. The main difference in these schemes is that the previous learns the phrase embeddings while the latter just merges already learned word embeddings to make the phrase embeddings. Similarly, \cite{fasttext} introduced an extension of the Skip-gram model \cite{w2v} that composes sub-word embeddings to make word embeddings with summation as the composition function. 

\section{The SPhrase model} \label{model}

The proposed model uses information about which words belong to which phrases. This information can be conveniently represented as simply the locations for where phrases start and end, hence the name, {\it Supervised Phrase Boundary Representations model} (SPhrase).
\par

The key assumption is that each word that comprises a phrase has the same context. This will produce an embedding where words that occur in the same phrase are likely to be close in the vector space. For example consider the sentence: \\
{\it  British airways to New York  has departed}.  \\
This sentence includes the (noun)  phrase `New York'. 
Following the procedure for Word2vec we focus on the target word `New'  using a context window of size 1. The target, context pairs are (New, to) and (New, York). Repeating this procedure for the target word `York', yields the target, context pairs (York, New) and (York, has).
\par
\begin{figure}[ht!]
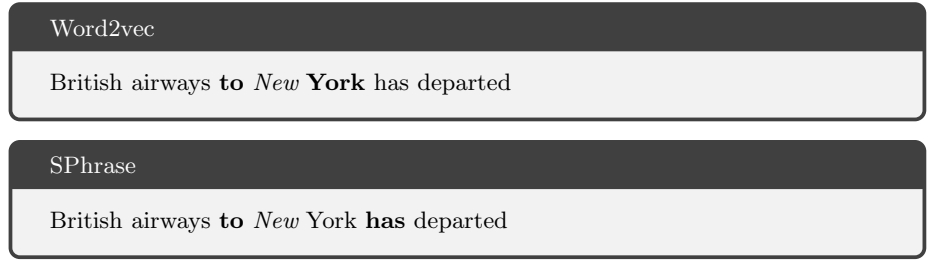

\begin{tcolorbox}[adjusted title=Word2vec ]
British airways {\bf to} {\it New} {\bf York} has departed
\end{tcolorbox}
\begin{tcolorbox}[adjusted title=SPhrase]
British airways {\bf to} {\it New} York {\bf has} departed
\end{tcolorbox}
\caption{Context words for the target word {\it New} using Word2vec and SPhrase. The context words are in bold.The context size is 1.} \label{fig:context}
\end{figure}

For SPhrase, the context differs from Word2vec, both target words in `New York' will have the same context based on the  words immediately surrounding the phrase, hence the SPhrase target context pairs are  (New, to), (New, has), (York, to), (York, has). Figure \ref{fig:context} highlights the context words for the word `New' for both Word2vec and SPhrase.

\par

In the above, we demonstrated the target context pairs induced by a target word that is a member of a phrase, where its context are individual words. In the following, we generalise the approach to handle the situation where phrases are part of a context. We do this by introducing the concept of a {\it unit}, where a unit consist of a sequence of words. A unit of length 1 represents individual words, a unit of length 2 represents two word phrases and so on for larger phrases.

\begin{figure}[ht!]
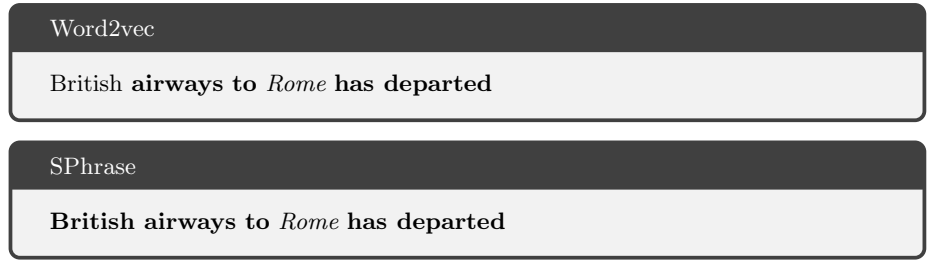

\begin{tcolorbox}[adjusted title=Word2vec ]
 British {\bf airways to} {\it Rome}  {\bf has departed}
\end{tcolorbox}
\begin{tcolorbox}[adjusted title=SPhrase]
{\bf British airways to} {\it Rome} {\bf has departed}
\end{tcolorbox}
\caption{Context words for the target word {\it Rome} using Word2vec and SPhrase. The context words are in bold. The context size is 2.} \label{fig:contextUnit}
\end{figure}

Thus we measure the context simply in terms of units. Figure \ref{fig:contextUnit} provides an example of a context of size 2 each side. Note that the left context for SPhrase contains 3 words. Thus the context size measured in words will be larger for SPhrase than Word2vec if there is a phrase within the context window. 

\subsection{SPhrase Context sampling}

A standard approach to reduce the computation involved in generating embeddings is to shorten the effective context length by using only a sample of words from a context \cite{w2v}. For SPhrase this can be achieved in several ways. First it can be done at the level of units not words, this is denoted {\it unit context sampling} (SPhrase). Second {\it random word context sampling} (R)\footnote{\scriptsize Pretrained embeddings are available at: https://github.com/ManniSingh/SPhrase} involves first performing unit context sampling, then for each unit that has a length greater than one only one word is sampled uniformly at random. This yields an effective context length that matches the context length of Word2vec. In addition to that, we generate embeddings named {\it without unit context sampling} (NU) where the target still is a unit but the context comprises individual words. 

\section{Methods and Datasets} \label{implimementation}

\subsection{Dataset}
In order to generate an embedding using our approach, we require a corpus that has phrases annotated.  Unfortunately this is not readily available, so we use a proxy for phrase annotation. In datasets that include hyperlinks we assume that the {\it hyperlink displayed text} is a phrase. One such data set is Wikipedia; we use the English Wikipedia dump version 20180920 that contains over 3 billion tokens. The proportion of tokens in phrases of length 2 is 2.5\%; of length 3,4,5, and greater is respectively 0.8\%, 0.3\%, 0.2\%, and less than 0.1\%. Obviously not all phrases are represented as hyperlink text and not all hyperlink texts are phrases. Indeed the longest  hyperlink text in our data set is of length 16,382 (it included internal formatting of Wikipedia). For our study we restricted maximum length to 10. The embedding vocabulary contained tokens with a frequency of at least 100 which gave us a total of 400,919 distinct tokens.

\subsection{Parameter settings}
Training is performed in mini-batches of 60,000 tokens per batch with candidate sampling of 5000 classes per batch (value dictated by the available computational resource). The remaining parameters use standard values, the learning rate is initialised to 0.001 and optimisation is  based on {\em Adam} optimiser \cite{adam} for stochastic learning. The learning decay is set to 10\% (i.e. learning rate * 0.9) after each epoch. The total number of the epochs is set to 20. The weighting scheme for selecting words in the context sampling is the same as for Word2vec \cite{w2v} 

\section{Evaluation}

There are two types of evaluation tasks commonly accepted: intrinsic and extrinsic.  
Intrinsic evaluation tasks determine the quality of embeddings. Under this class, word similarity/relatedness tasks are generally based on cosine distance as a metric to find similarity between two word vectors. Extrinsic evaluation tasks, on the other hand, are based on specific downstream tasks such as, named entity recognition (NER), sentiment classification, topic detection. In this work, we are doing similarity based intrinsic evaluation and NER based extrinsic evaluation.

\section{Experimental design} \label{experiment}

\subsection{Intrinsic evaluation}
The following experiments fit into the so-called {\it intrinsic} category of embedding evaluation. We aim to demonstrate that although the total number of phrases in our dataset is small compared to the number of words, they do have a positive impact on the resulting embeddings. In order to determine an optimal configuration of the method, intrinsic evaluation is done on embeddings trained on the first 10\% of the corpus; see Figure \ref{fig:task2}, As a result, the extrinsic evaluation described Section \ref{extrinsic}, the performance of the optimal configuration in this evaluations is: SPhrase (R) with window size 5. For the extrinsic evaluation only the optimal configuration is used and the embeddings are trained on the full corpus.    

In the following experiments we compare SPhrase embeddings with the ones generated by Word2vec. It is known that increasing the context window size generally improves the quality of the embedding. Recall that the expected context size for each target word is the same for Word2vec and SPhrase due to word context sampling. 

We expect that words in phrases should be mapped to similar locations in the embedding, i.e. words within a phrase should be closer together than words that are not in the same phrase. In the following we first perform experiment on pairwise similarity and then we investigate further structure with an analogy task. 

\subsubsection{ Pairwise Similarity}
For pairwise similarity experiments we use phrases from three datasets.

\begin{itemize}
    \item CoNLL-2003 English  dataset \cite{conll2003}. From this dataset multi-word named entities were extracted. These are used as phrases, in total there are 12,999. The maximum phrase length is 7 in this dataset, so we restricted the following two datasets to this as well. 
    \item From our Wikipedia training corpus we obtained 16,470 phrases from the first 1,000000 tokens. This dataset comes from our training data, so we assume we should obtain good results in this case.
    \item Bristol \cite{bristol}-  from this dataset we selectively used the entity list and found 87,209 phrases.  
\end{itemize}

\noindent
\begin{figure}[!t]
\rotatebox{90}{Score}
\begin{subfigure}{0.28\textwidth}
  \caption*{\hspace{0.5cm}\bf{window size: 3 }}
  \includegraphics[scale=0.28,angle=270]{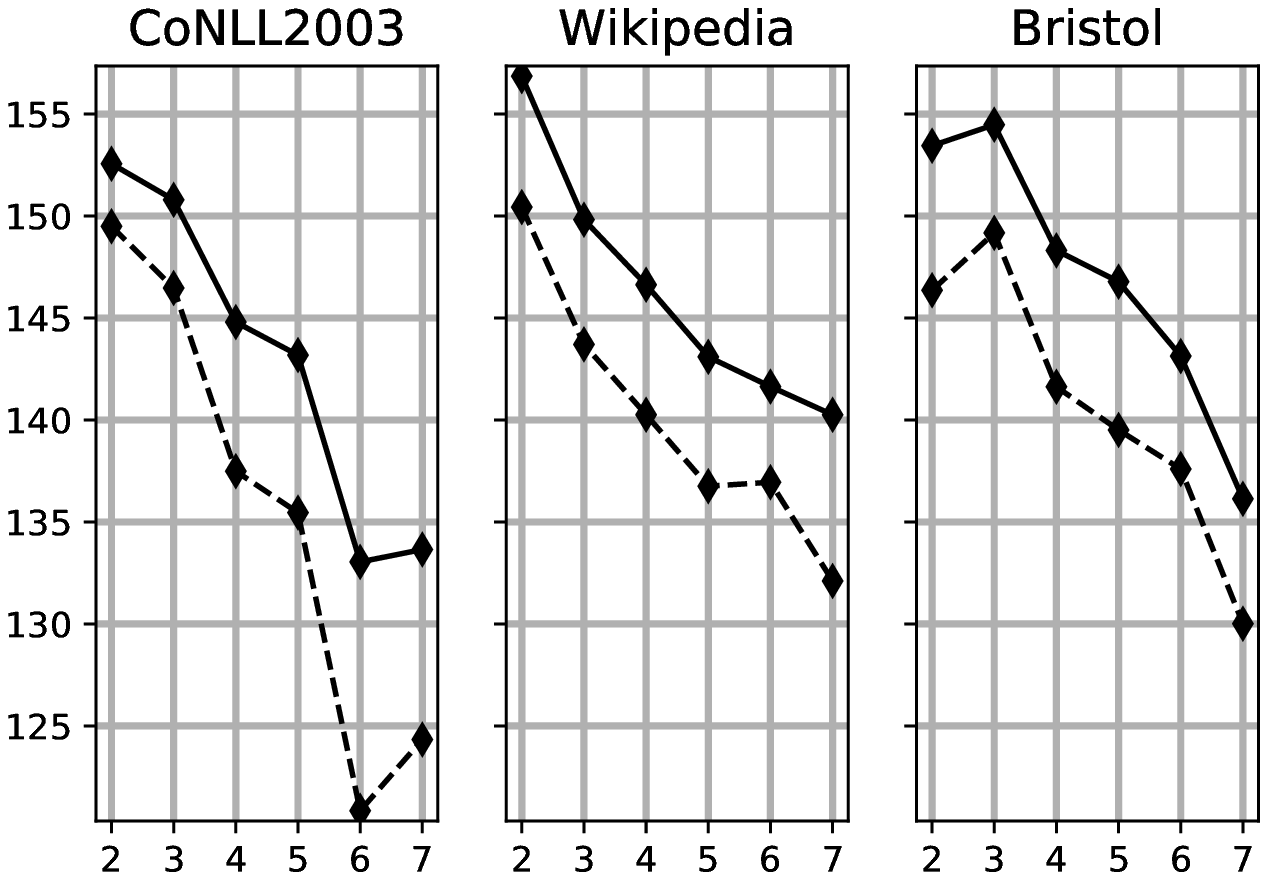}
  \caption*{\hspace{0.5cm} SPhrase}
\end{subfigure}\hfil 
\begin{subfigure}{0.28\textwidth}
  \caption*{\hspace{0.5cm}\bf{ window size: 5 }}
  \includegraphics[scale=0.28,angle=270]{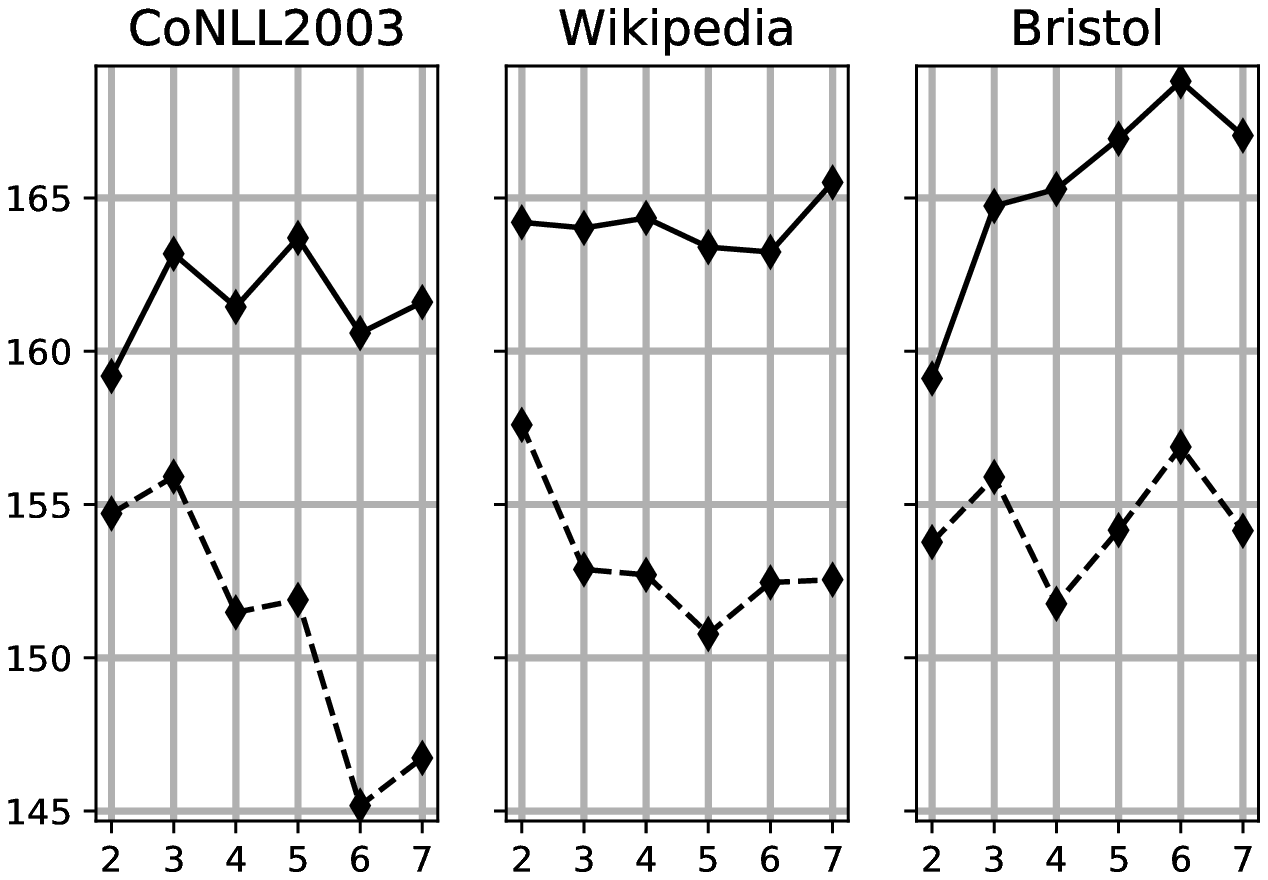}
  \caption*{\hspace{0.5cm} SPhrase}
\end{subfigure}\hfil 
\begin{subfigure}{0.28\textwidth}
  \caption*{\hspace{0.5cm}\bf{ window size: 10 }}
  \includegraphics[scale=0.28,angle=270]{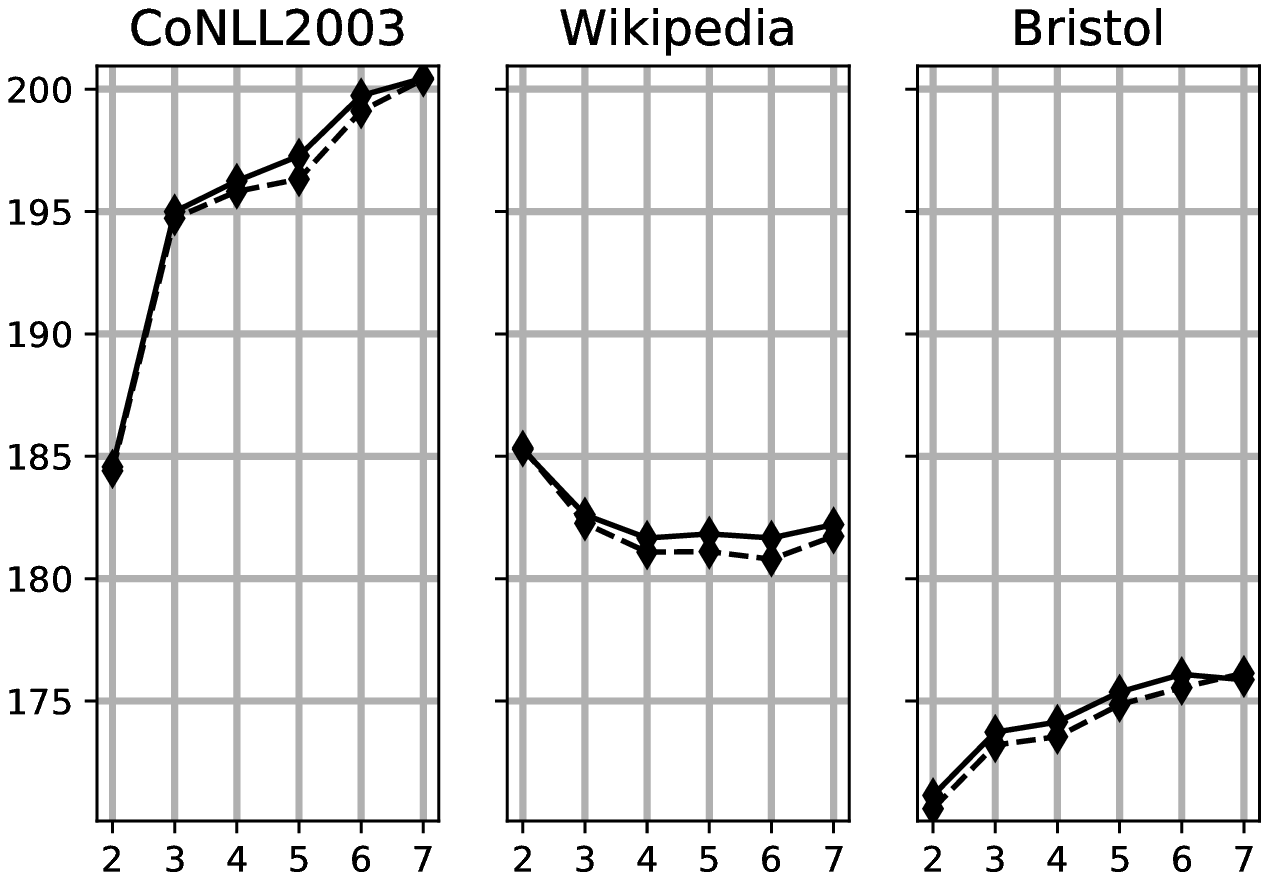}
  \caption*{\hspace{0.5cm} SPhrase}
\end{subfigure}

\medskip
\rotatebox{90}{Score}
\begin{subfigure}{0.28\textwidth}
  \includegraphics[scale=0.28,angle=270]{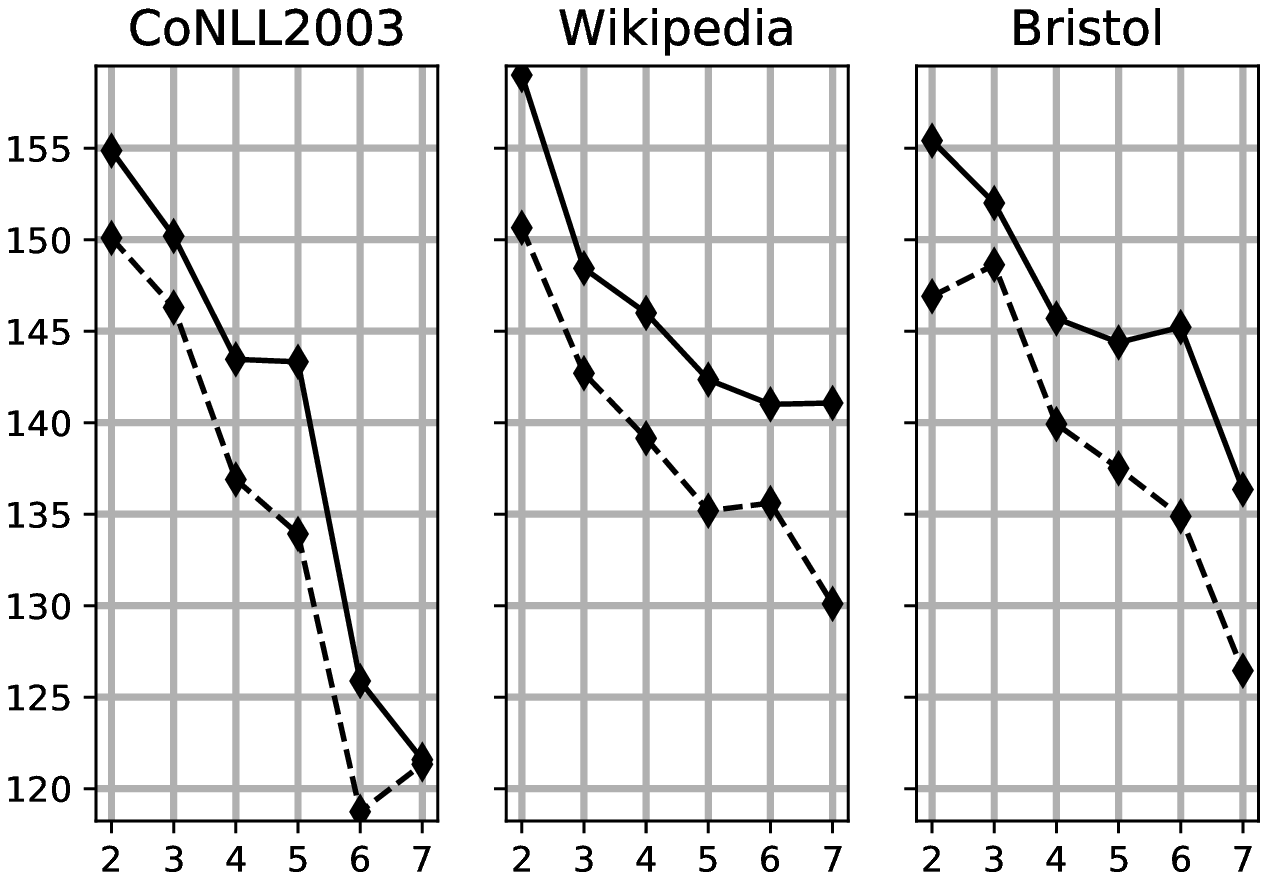}
  \caption*{\hspace{0.5cm} SPhrase (NU)}
\end{subfigure}\hfil 
\begin{subfigure}{0.28\textwidth}
  \includegraphics[scale=0.28,angle=270]{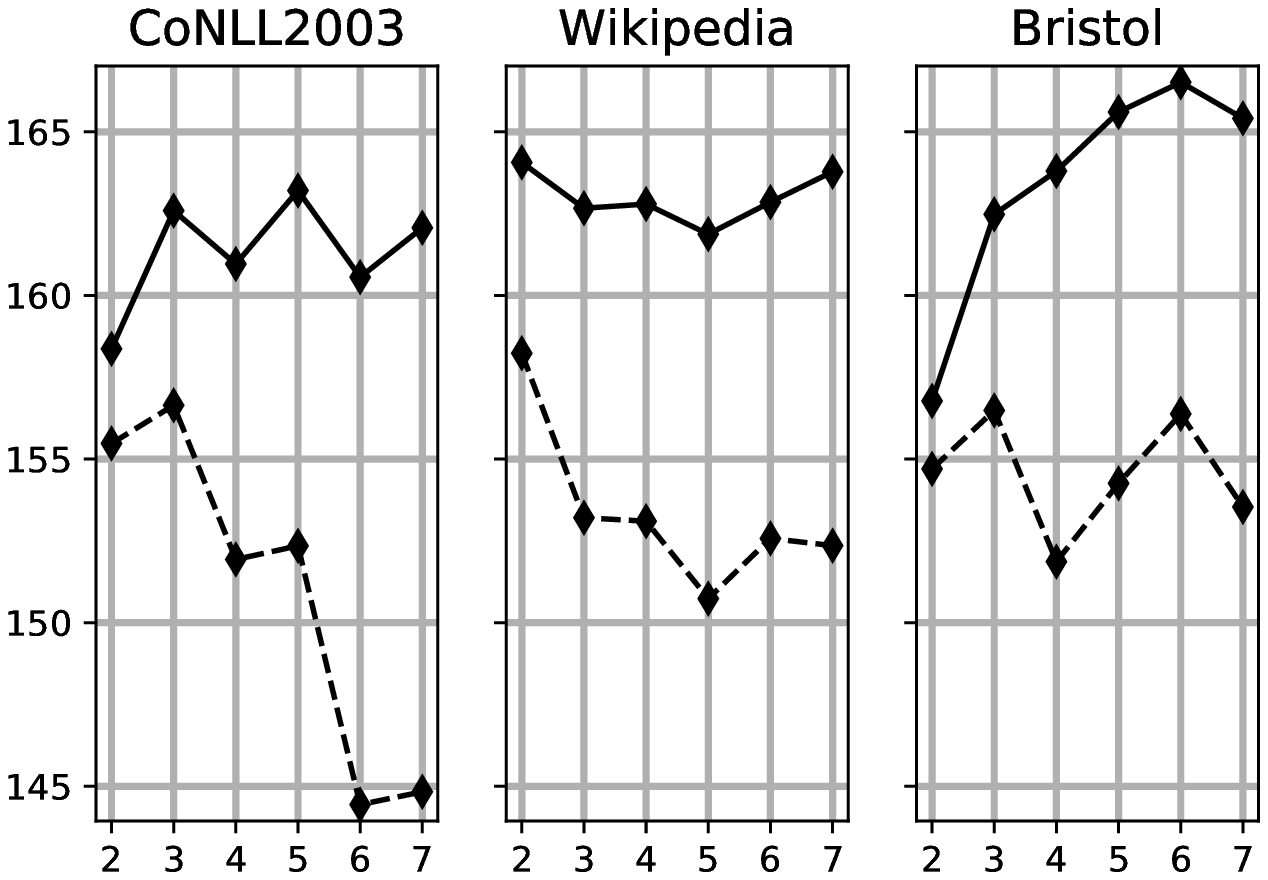}
  \caption*{\hspace{0.5cm} SPhrase (NU)}
\end{subfigure}\hfil 
\begin{subfigure}{0.28\textwidth}
  \includegraphics[scale=0.28,angle=270]{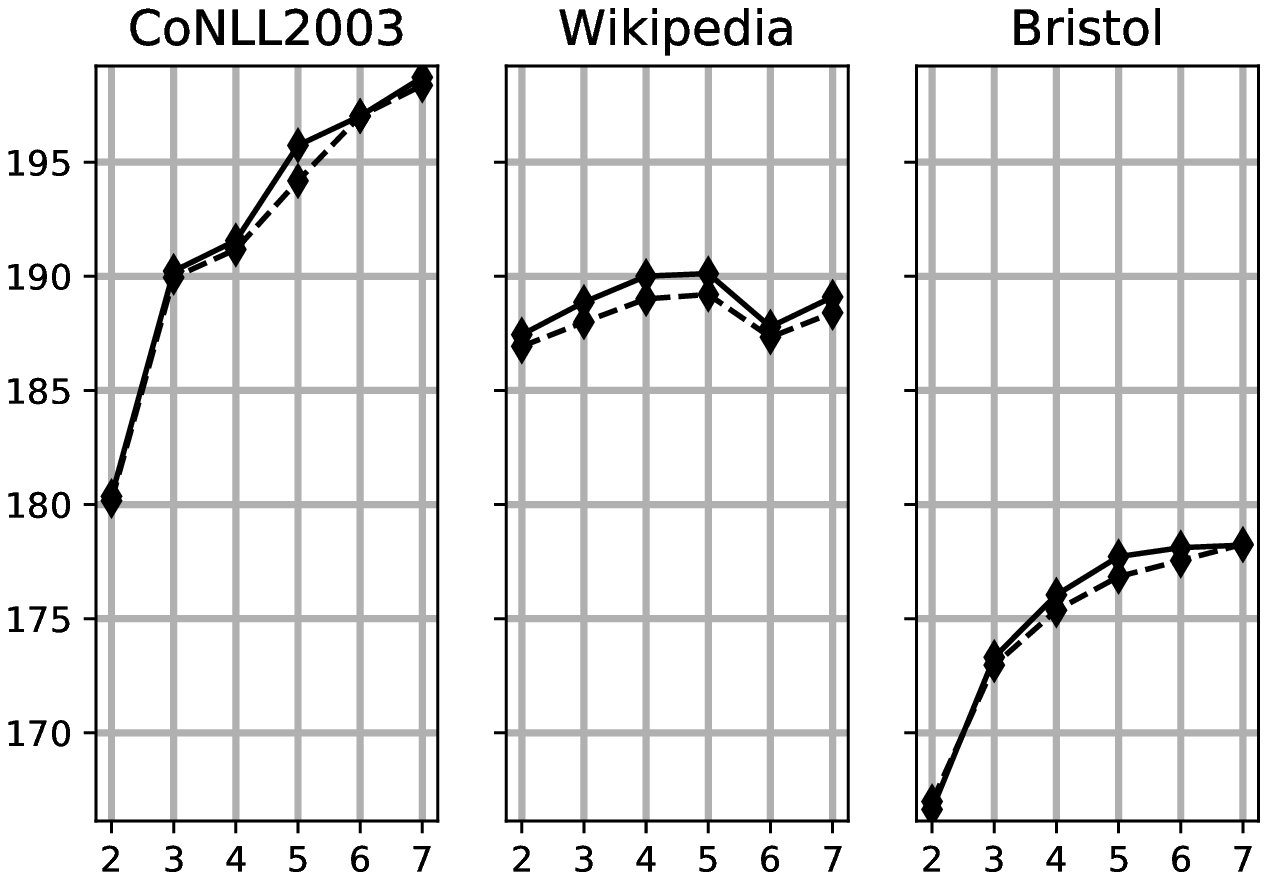}
  \caption*{\hspace{0.5cm} SPhrase (NU)}
\end{subfigure}

\medskip
\rotatebox{90}{Score}
\begin{subfigure}{0.28\textwidth}
  \includegraphics[scale=0.28,angle=270]{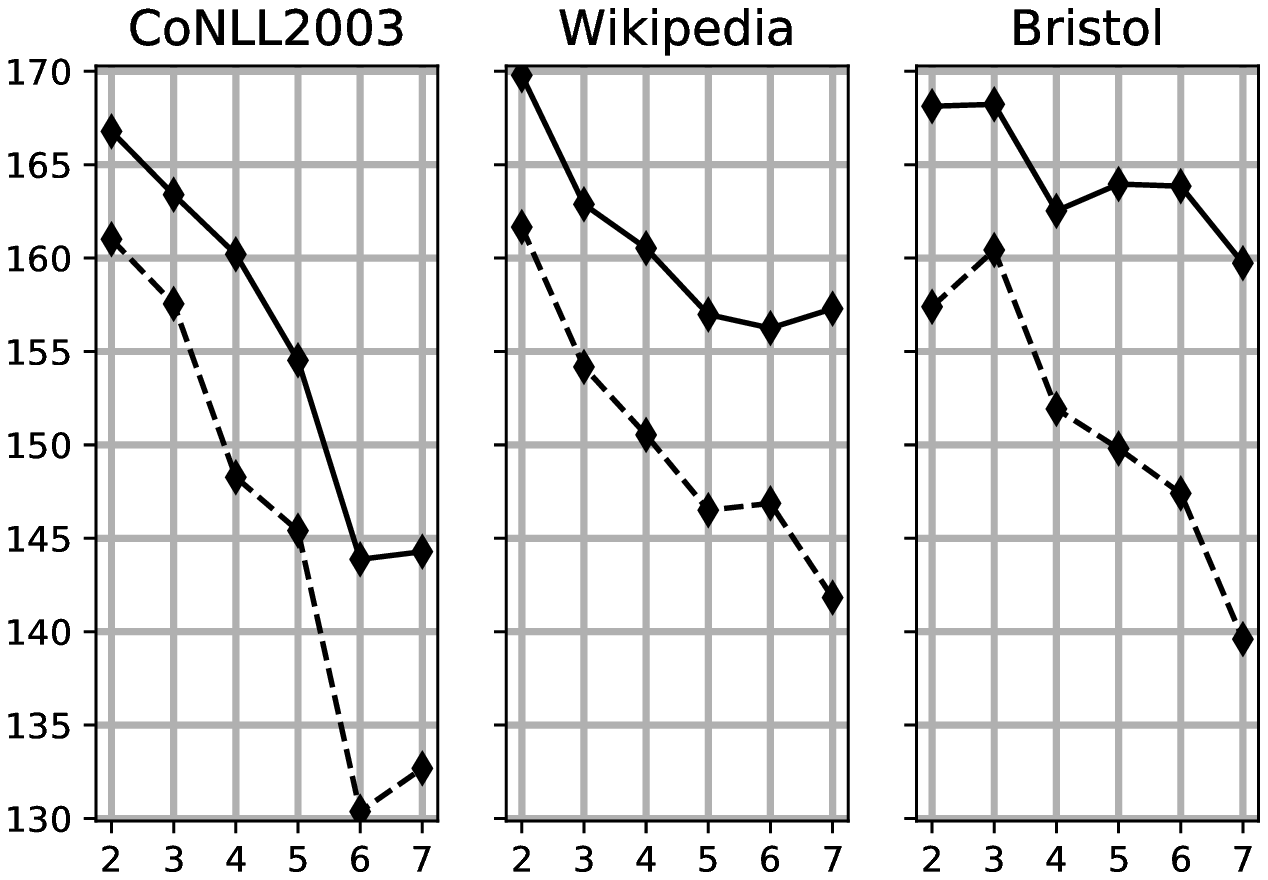}
  \caption*{\hspace{0.5cm} SPhrase (R)}
\end{subfigure}\hfil 
\begin{subfigure}{0.28\textwidth}
  \includegraphics[scale=0.28,angle=270]{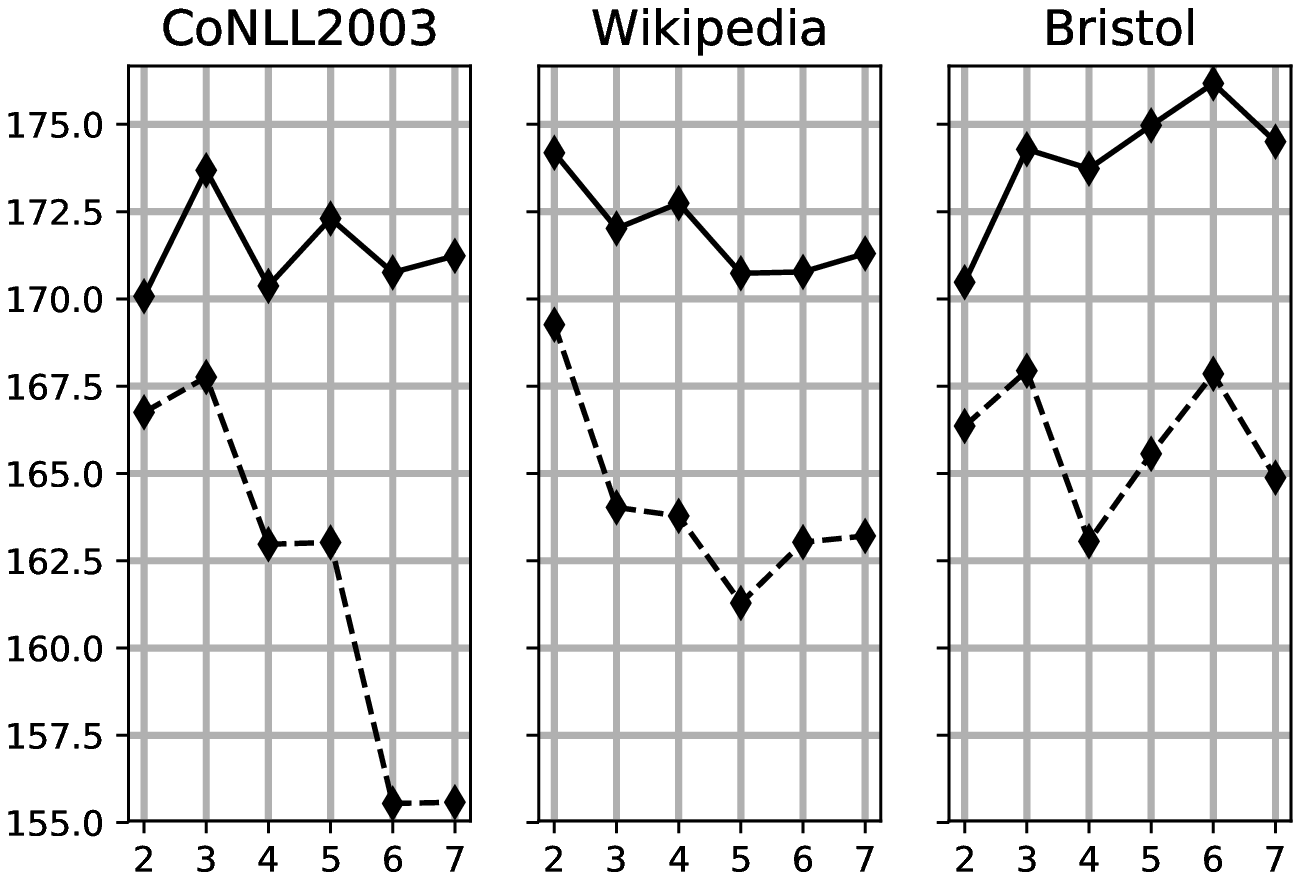}
  \caption*{\hspace{0.5cm} SPhrase (R)}
\end{subfigure}\hfil 
\begin{subfigure}{0.28\textwidth}
  \includegraphics[scale=0.28,angle=270]{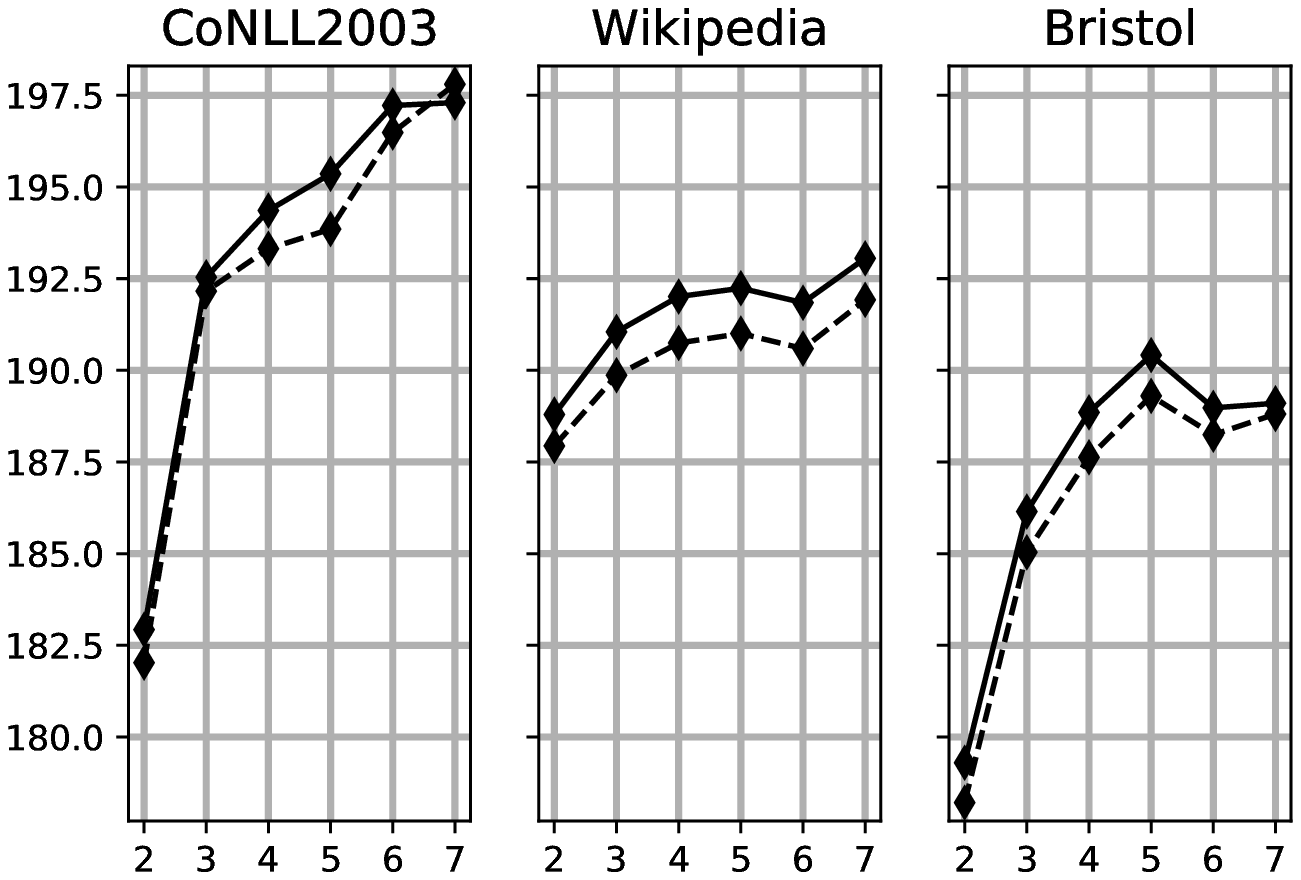}
  \caption*{\hspace{0.5cm} SPhrase (R)}
\end{subfigure}
\caption{Similarity scores comparison for the phrases relative to 100 random words representing: \textit{unit context sampling} (SPhrase), Without \textit{unit context sampling} (NU) and, with \textit{random word context sampling} (R). Where SPhrase (in bold) and Word2vec (dashed) are compared on phrase lengths 2-7 (in horizontal axis) with higher the score the better it performed. }
\label{fig:task2}
\end{figure}

\noindent
In order to investigate how the distances of words within a phrase compare to distances of words with random words in the datasets we use the following,

\begin{gather*}
\text{Similarity Score } = \dfrac{1}{N_l (l-1)}{\sum_{i=1}^{l-1} b(w_i,w_{i+1},r)}\\
\shortintertext{where,}
b(w_i,w_{i+1},r) = \left\{ \begin{array}{ll} 
                1 & \hspace{5mm} s(w_i,w_{i+1})> s(w_i,r), \\
                0 & \hspace{5mm} \text{otherwise,} \\
                \end{array} \right.
\end{gather*}

where $r$ is a word selected at random from another phrase. A new word is drawn for each phrase pair comparison. The similarity score is calculated 100 times and the overall average is taken in order to reduce the noise generated by selecting only one word for each comparison. The interpretation of this is similar to the cosine score in that the larger the value the better.

\noindent
We computed scores for phrase lengths up to and including length 7. We have used context window sizes 3, 5 and 10. Figure \ref{fig:task2} shows these scores for the context sampling regimes: with \textit{unit context sampling}, without \textit{unit context sampling}, and \textit{word context sampling}.  

We can see that regardless of the embedding, the scores in general reduce as the phrase gets longer. However, the larger the window size the more Word2vec and SPhrase agree. This is what we should expect, since there will be greater overlap in the context words between SPhrase and Word2vec. Nevertheless we see that, overall, SPhrase performs better.

\subsubsection{Google Analogy Test set}

Analogy based tasks are widely used, e.g. \cite{semeval02,Bruni,Finkelstein} to evaluate the quality of word embeddings. One well known test set is the  Google analogy test set \cite{w2v}. This dataset comprises rows of four words, such as \verb!known unknown informed uninformed!. The analogy task is to predict the final word using the first three using simple vector addition/subtraction of their vector representations. Informally the task attempts to show how well words follow the vector relationship \\  {\it unknown - known = uninformed - informed}\\

\begin{table}[h]
\caption{Scores on Google analogy dataset with \textit{unit context sampling} (SPhrase), here accuracy is the total correct count on the total count of instances.}
\resizebox{\textwidth}{!}{%
\begin{tabular}{|l|N{3}{3}|N{3}{3}|N{3}{3}|N{3}{3}|N{3}{3}|N{3}{3}|r|}
\hline
& \multicolumn{6}{c|}{Accuracy - displayed to 3 decimal places} &\\
\hline
& \multicolumn{2}{c||}{Window size 3} & \multicolumn{2}{c||}{Window size 5} & \multicolumn{2}{c|}{Window size 10} &\\
\hline
	&	SPhrase	&	{Word2vec}	&	SPhrase	&	{Word2vec}	&	SPhrase	&	{Word2vec}	&	Count	\\
\hline
\textit{capital-world}	&	0.727454	&	0.627542	&	0.746463	&	0.658488	&	0.814766	&	0.78183	&	4524	\\
\textit{capital-common-countries}	&	0.871542	&	0.847826	&	0.940711	&	0.855731	&	0.976285	&	0.940711	&	506	\\
\textit{city-in-state}	&	0.659911	&	0.47953	&	0.715039	&	0.582894	&	0.646534	&	0.676936	&	2467	\\
gram3-comparative	&	0.848348	&	0.805556	&	0.757508	&	0.813063	&	0.642643	&	0.66967	&	1332	\\
gram2-opposite	&	0.222906	&	0.220443	&	0.220443	&	0.221675	&	0.205665	&	0.204433	&	812	\\
gram8-plural	&	0.754505	&	0.735736	&	0.714715	&	0.743994	&	0.641141	&	0.727477	&	1332	\\
gram4-superlative	&	0.378788	&	0.395722	&	0.34492	&	0.36631	&	0.278966	&	0.262032	&	1122	\\
gram9-plural-verbs	&	0.63908	&	0.558621	&	0.535632	&	0.545977	&	0.452874	&	0.52069	&	870	\\
gram6-nationality-adjective	&	0.846154	&	0.783615	&	0.838024	&	0.814884	&	0.854284	&	0.853033	&	1599	\\
family	&	0.602767	&	0.594862	&	0.594862	&	0.63834	&	0.581028	&	0.543478	&	506	\\
gram7-past-tense	&	0.471795	&	0.515385	&	0.473718	&	0.492308	&	0.441026	&	0.469872	&	1560	\\
currency	&	0.047344	&	0.04157	&	0.020785	&	0.020785	&	0.018476	&	0.016166	&	866	\\
gram1-adjective-to-adverb	&	0.103831	&	0.086694	&	0.118952	&	0.120968	&	0.132056	&	0.148185	&	992	\\
gram5-present-participle	&	0.517045	&	0.519886	&	0.50947	&	0.485795	&	0.479167	&	0.454545	&	1056	\\
\hline
all	&	0.600747	&	0.544515	&	0.596807	&	0.564623	&	0.581253	&	0.586881	&	19544	\\
\hline
\end{tabular}
}
\label{tab:google}
\end{table}

\begin{table}[h!]
\caption{Scores on Google analogy dataset without \textit{unit context sampling} (NU), here accuracy is the total correct count on the total count of instances.}
\begin{center}
\resizebox{\columnwidth}{!}{%
\begin{tabular}{|l|N{3}{3}|N{3}{3}|N{3}{3}|N{3}{3}|N{3}{3}|N{3}{3}|r|}
\hline
& \multicolumn{6}{c|}{Accuracy - displayed to 3 decimal places} &\\
\hline
& \multicolumn{2}{c||}{Window size 3} & \multicolumn{2}{c||}{Window size 5} & \multicolumn{2}{c|}{Window size 10} &\\
\hline
	&	SPhrase	&	{Word2vec}	&	SPhrase	&	{Word2vec}	&	SPhrase	&	{Word2vec}	&	Count	\\
\hline

\textit{capital-world}	&	0.671309	&	0.627542	&	0.72458	&	0.658488	&	0.74359	&	0.78183	&	4524	\\
\textit{capital-common-countries}	&	0.881423	&	0.847826	&	0.934783	&	0.855731	&	0.928854	&	0.940711	&	506	\\
\textit{city-in-state}	&	0.652615	&	0.47953	&	0.644913	&	0.582894	&	0.652209	&	0.676936	&	2467	\\
gram3-comparative	&	0.705706	&	0.805556	&	0.695946	&	0.813063	&	0.518769	&	0.66967	&	1332	\\
gram2-opposite	&	0.216749	&	0.220443	&	0.197044	&	0.221675	&	0.172414	&	0.204433	&	812	\\
gram8-plural	&	0.725976	&	0.735736	&	0.712462	&	0.743994	&	0.660661	&	0.727477	&	1332	\\
gram4-superlative	&	0.272727	&	0.395722	&	0.297683	&	0.36631	&	0.269162	&	0.262032	&	1122	\\
gram9-plural-verbs	&	0.577011	&	0.558621	&	0.548276	&	0.545977	&	0.477011	&	0.52069	&	870	\\
gram6-nationality-adjective	&	0.854909	&	0.783615	&	0.821138	&	0.814884	&	0.826767	&	0.853033	&	1599	\\
family	&	0.56917	&	0.594862	&	0.55336	&	0.63834	&	0.501976	&	0.543478	&	506	\\
gram7-past-tense	&	0.453205	&	0.515385	&	0.483333	&	0.492308	&	0.414103	&	0.469872	&	1560	\\
currency	&	0.039261	&	0.04157	&	0.024249	&	0.020785	&	0.027714	&	0.016166	&	866	\\
gram1-adjective-to-adverb	&	0.13004	&	0.086694	&	0.173387	&	0.120968	&	0.168347	&	0.148185	&	992	\\
gram5-present-participle	&	0.511364	&	0.519886	&	0.50947	&	0.485795	&	0.492424	&	0.454545	&	1056	\\
\hline
all	&	0.56534	&	0.544515	&	0.576494	&	0.564623	&	0.552804	&	0.586881	&	19544	\\
\hline
\end{tabular}
}
\label{tab:google_nc}
\end{center}
\end{table}

The dataset is divided into categories, some of which are inherently phrase-based. In the category \verb!capital-common-countries! a typical line is:\\
\verb!Athens Greece Baghdad Iraq!\\
Both {\it Athens Greece} and  {\it Baghdad Iraq} can be reasonably construed to be  phrases, unlike in the first example above. Two other categories have this same character, namely
\verb!capital-world! and \verb!city-in-state!. \\
Example rows are: 
\verb!Athens Greece Canberra Australia!
and \\
\verb!Chicago Illinois Houston Texas! respectively.

With this in mind we show the accuracy of SPhrase and Word2vec stratified by category, in addition to the overall accuracy that is usually reported. The categories that have a phrasal quality are italicised in Tables 1-3. We see that, overall, SPhrase performs better in these categories.

\begin{table}
\caption{Scores on Google analogy dataset with \textit{random word context sampling} (R), here accuracy is the total correct count on the total count of instances.}
\begin{center}
\resizebox{\textwidth}{!}{%
\begin{tabular}{|l|N{3}{3}|N{3}{3}|N{3}{3}|N{3}{3}|N{3}{3}|N{3}{3}|r|}
\hline
& \multicolumn{6}{c|}{Accuracy - displayed to 3 decimal places} &\\
\hline
& \multicolumn{2}{c||}{Window size 3} & \multicolumn{2}{c||}{Window size 5} & \multicolumn{2}{c|}{Window size 10} &\\
\hline
	&	SPhrase	&	{Word2vec}	&	SPhrase	&	{Word2vec}	&	SPhrase	&	{Word2vec}	&	Count	\\
\hline

\textit{capital-world}	&	0.637268	&	0.627542	&	0.718391	&	0.658488	&	0.765915	&	0.78183	&	4524	\\
\textit{capital-common-countries}	&	0.857708	&	0.847826	&	0.903162	&	0.855731	&	0.952569	&	0.940711	&	506	\\
\textit{city-in-state}	&	0.663559	&	0.47953	&	0.623429	&	0.582894	&	0.662748	&	0.676936	&	2467	\\
gram3-comparative	&	0.844595	&	0.805556	&	0.802553	&	0.813063	&	0.682432	&	0.66967	&	1332	\\
gram2-opposite	&	0.224138	&	0.220443	&	0.245074	&	0.221675	&	0.195813	&	0.204433	&	812	\\
gram8-plural	&	0.771772	&	0.735736	&	0.731231	&	0.743994	&	0.655405	&	0.727477	&	1332	\\
gram4-superlative	&	0.37344	&	0.395722	&	0.392157	&	0.36631	&	0.256684	&	0.262032	&	1122	\\
gram9-plural-verbs	&	0.574713	&	0.558621	&	0.586207	&	0.545977	&	0.473563	&	0.52069	&	870	\\
gram6-nationality-adjective	&	0.818011	&	0.783615	&	0.82364	&	0.814884	&	0.831144	&	0.853033	&	1599	\\
family	&	0.614625	&	0.594862	&	0.581028	&	0.63834	&	0.594862	&	0.543478	&	506	\\
gram7-past-tense	&	0.478846	&	0.515385	&	0.519872	&	0.492308	&	0.459615	&	0.469872	&	1560	\\
currency	&	0.040416	&	0.04157	&	0.024249	&	0.020785	&	0.023095	&	0.016166	&	866	\\
gram1-adjective-to-adverb	&	0.089718	&	0.086694	&	0.127016	&	0.120968	&	0.172379	&	0.148185	&	992	\\
gram5-present-participle	&	0.525568	&	0.519886	&	0.454545	&	0.485795	&	0.479167	&	0.454545	&	1056	\\
\hline
all	&	0.575778	&	0.544515	&	0.5877	&	0.564623	&	0.576494	&	0.586881	&	19544	\\
\hline
\end{tabular}
}

\label{tab:google_rand}
\end{center}
\end{table}

\subsection{Extrinsic evaluation} \label{extrinsic}

\begin{table}[h]
    \centering
    \caption{Comparison of Word2vec with SPhrase(NU) on Conll2003 English and Wikigold dataset}
    \begin{tabular}{|c|c|c|}
        \hline
        Model & Conll2003Eng & Wikigold  \\
        \hline
         Word2Vec & $83.82\pm0.3831$ & $55.49\pm0.4708$ \\
         SPhrase &  $\mathbf{88.93\pm0.1115}$ & $\mathbf{66.01\pm0.4172}$ \\
         \hline
    \end{tabular}
    \label{tab:NER}
\end{table}

We use Conll2003 English \cite{conll2003} and Wikigold \cite{wikigold} to evaluate the performance of the embeddings generated. The Conll dataset is widely used to evaluate various NER based models. It contains 203,621 tokens in the training set, while validation and test set contains 51,362 and 46,435 tokens respectively. On the other hand, Wikigold provides a single data file of 39,007 tokens that we used for testing while the NER models were trained with Conll train and validation data. We used SPhrase (R) model with window size 5 since this configuration demonstrated significant improvements over Word2vec as shown in Figure \ref{fig:task2}. We recreated the BLSTMs and CRF based model \cite{lample2016neural} but without any feature engineering. We trained this in 20 epochs with evaluating on validation data each time. We performed 10 instances for each of these models and presented the range of F1 scores (using Conll2003 evaluation script). Table \ref{tab:NER}  displays the results that show a significant improvement over the Word2vec model trained on the same corpus.  

\section{Concluding remarks}

This investigation demonstrates that using phrasal information can directly enrich word embeddings. In this work, we presented an alternative context sampling technique to that used in skip-gram Word2vec. We note that the SPhrase approach is not limited to augmenting Word2Vec, it can also be applied to  morphological extensions such as Fasttext \cite{fasttext}.

We used the displayed text from hyperlinks as a proxy for phrases, and in this sense SPhrase is supervised. We are, however, planning to generalise the methodology by investigating whether we can identify useful phrase boundaries in a completely  unsupervised fashion.

\bibliography{references}
\bibliographystyle{splncs04}

%
%
%
%

\end{document}